\documentclass[letterpaper, 10 pt, conference]{ieeeconf}  

\IEEEoverridecommandlockouts                              

\usepackage{graphicx} 
\usepackage{gensymb}
\usepackage{hyperref}
\hypersetup{
    colorlinks=true,
    linkcolor=black,
    filecolor=magenta,      
    urlcolor=black,
}
\usepackage{graphics} 
\usepackage{epsfig} 
\usepackage{mathptmx} 
\usepackage{times} 
\usepackage{amsmath} 
\usepackage{amssymb}  

\usepackage{subfigure}
\usepackage{tabularx}
\usepackage{dblfloatfix}
\usepackage{textcomp}
\usepackage{hyperref}
\usepackage{color}
\newcommand{\RNum}[1]{\uppercase\expandafter{\romannumeral #1\relax}}

\title{\LARGE \bf
Improved GelSight Tactile Sensor for Measuring Geometry and Slip
}

\author{Siyuan Dong$^{1}$, Wenzhen Yuan$^{2}$ and Edward H. Adelson$^{3}$
\thanks{$^{1}$Department of Electrical Engineering and Computer Science, and Computer Science and Artificial Intelligence Laboratory (CSAIL), MIT, Cambridge, MA 02139, USA     {\tt\small sydong@mit.edu}}%
\thanks{$^{2}$Department of Mechanical Engineering, and CSAIL, MIT, Cambridge, MA 02139,USA 
        {\tt\small yuan\_wz@csail.mit.edu}}%
\thanks{$^{3}$Department of Brain and Cognitive Sciences, and CSAIL, MIT, Cambridge, MA 02139, USA     {\tt\small adelson@csail.mit.edu}}%
}

\begin{document}

\maketitle
\thispagestyle{empty}
\pagestyle{empty}

\begin{abstract}
A GelSight sensor uses an elastomeric slab covered with a reflective membrane to measure tactile signals. It measures the 3D geometry and contact force information with high spacial resolution, and successfully helped many challenging robot tasks. A previous sensor~\cite{GelSightUSB}, based on a semi-specular membrane, produces high resolution but with limited geometry accuracy. In this paper, we describe a new design of GelSight for robot gripper, using a Lambertian membrane and new illumination system, which gives greatly improved geometric accuracy while retaining the compact size. We demonstrate its use in measuring surface normals and reconstructing height maps using photometric stereo. We also use it for the task of slip detection, using a combination of information about relative motions on the membrane surface and the shear distortions. Using a robotic arm and a set of 37 everyday objects with varied properties, we find that the sensor can detect translational and rotational slip in general cases, and can be used to improve the stability of the grasp.
\end{abstract}

\section{INTRODUCTION}

Tactile sensing is an important way for robots to sense and interact with the environment. With a tactile sensor in its hand, a robot can know whether it is holding some object or whether the gripping force is proper. Examples of tactile sensors designed in the past decades can be found in~\cite{TactileReview2010, TactileSensorReview2011, KAPPASSOV2015195}.


Among many robotic tasks that require assistance of tactile sensing, the most important task is to detect whether the robot has safely grasped an object. Slip, a common grasp failure, will occur when the gripping force is not large enough. The warning signals of slipping objects, such as the stretch of fingertip skin and the subtle vibration of a sliding object, can be easily perceived by human. For a long time, researchers have been trying to develop tactile sensors capable of detecting slip~\cite{slipreview2013}.  Tactile sensors with this capability measures various tactile signals, including contact force, vibration, acceleration, and stretch of the sensor surface. Recently, Su~\textit{et al}. ~\cite{biotachslip2015} and Ajoudani~\textit{et al}. \cite{IIT2016slipcontrol} show a grasp control system with the slip-detection function enabled by vibration measurement. It has been shown that the system allow the robot to adjust the gripping force according to the detected slip condition and consequently execute a more stable grasp. 
However, engineering a slip detection device robust to the weight and geometry of the object is a challenging problem. 

\begin{figure}[thpb]
	\centering
	\includegraphics[scale=.3] {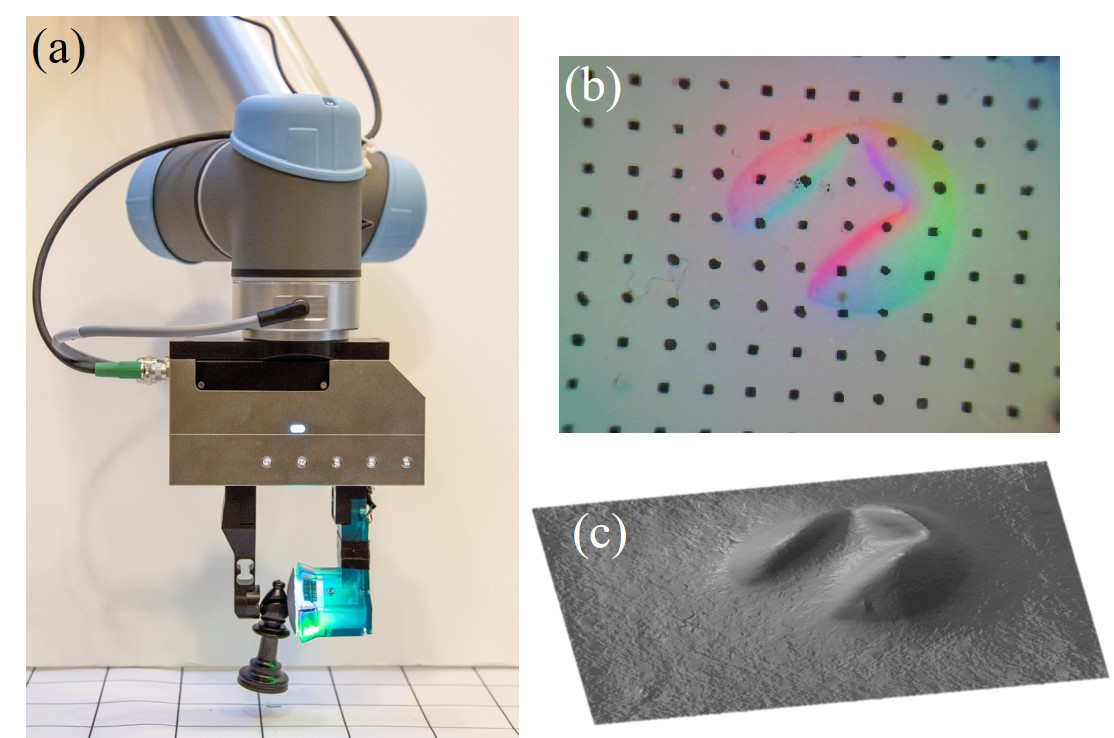}
	\caption{(a) A parallel gripper WSG-50 with the new GelSight sensor gripping a chess Bishop.(b) Image captured by GelSight sensor during the grasp (c) The reconstructed 3D geometry from (b)}
	\label{fig:sensor in gripper}
\end{figure}


An optical tactile sensor, GelSight~\cite{GelSight2009,GelSight2011}, was introduced to obtain a high-resolution tactile image of the 3D topography of the contact surface. The Fingertip GelSight, a scaled-down version of GelSight, was developed to be mounted on a robot gripper~\cite{GelSightUSB}. It provides high spatial resolution (640$\times$480 pixels, 0.024 mm/pixel) and has successfully assisted many challenging robotic tasks. However, Despite its successes in these robotic applications, the Fingertip GelSight sensor faces several challenges, such as unsatisfactory precision of its surface normal measurement and fabrication difficulty. 

To address these issues, we propose a new design for the Fingertip Gelsight sensor in this paper. The new sensor redesigns the illumination system to make the light more uniform and suitable for lambertian reflective surface and therefore improve the quality of the reconstructed 3D height map of the contact surface. Moreover, the entire frame of the new sensor can be easily printed by 3D printers, and the illumination system of the new design can be easily assembled. The detailed sensor fabrication process is provided in the sensor design section and the Solidworks files of the sensor frame are available online. Consequently, the new sensor, with a highly simplified and standardized fabrication process, is much easier to be reproduced. 


We test the new sensor with the task of measuring slip during grasp, and use the measurement to enhance grasp. 
We use the methods that are based on the fundamental definitions of slip -- the phenomenon of the relative displacement between the finger and the objects; and the general way how humans detect slip -- measuring the skin stretch of the soft fingers. They are all universal for varied objects. Therefore, the methods can be more generally applied, without requiring pre-knowledge of the objects' geometries, weights, materials or surface roughness. This measurement of general slip cases will help robots to better monitor the state of grasp, and increase the success rate of grasping unknown or more complicated objects. We believe the method will make intelligent robots more adaptive to the real environment.

This paper is structured as follows: in section \RNum{2}, we introduce the related work about optical based tactile sensors and general methods to detect slip. In section \RNum{3} and \RNum{4}, we describe the design of the new sensor and quantitatively evaluate its performance of geometry measurements. In section \RNum{5} and \RNum{6}, we introduce the slip detection method we are using and demonstrate the capability of the new sensor to detect slip. Finally, we summarize the the contribution of this paper and discuss potential applications. 

\section{RELATED WORK}

\subsection{Optical Tactile Sensors}
Optical tactile sensors convert signals of the contact deformation into images and thus achieve high spatial resolution and high sensitivity to the contact force. Optical tactile sensors capture the deformation of the contacting surface by a camera and infer the shape of the detected object, the shear force, the torque and other important information. In 1993, Jiar \textit{et al}.~\cite{jiar1993high} built a compact tactile sensing prototype for robotic manipulations, which was able to reconstruct the 2D shape of the detected object and roughly estimate the gripping force by capturing binary images with a CCD camera. Another 
finger shaped tactile sensor using optical waveguide, designed by MAEKAWA \textit{et al}.~\cite{maekawa1993finger}, was used to detect the contact point and normal to the surface. Later on, M Ohka \textit{et al}.~\cite{ohka1996data} proposed a tactile sensor that used camera to record the contacting area of a rubber pressed against an array of pyramidal projections. The three-dimensional force as well as the stiffness was reconstructed. In 2000, a compact tactile sensor complemented by Ferrier and Brockett~\cite{Ferrier2000}  successfully reconstructed the coarse 3D shape of the detected object. By using markers on the membrane and recording the flow field of the marker movement, they sketched the 3D map of the deformation. The idea of adding textures such as arrays of dots or grids on the contact surface of the tactile sensor was also implemented in~\cite{Bristol2009,sato2008measurement,nagata1999feature,GelForce2005} to encode edge information, reconstruct surface traction fields and force vector field. However, this method was not able to reconstruct 3D maps with high spatial resolution, since the texture on the surface was large and sparse. Therefore, The achieved spatial resolution was limited by the texture, which is much lower than that of the imaging system.

\subsection{GelSight Sensor}

The GelSight sensor is also an optical tactile sensor, which is mainly designed to achieve high precision for the measurement of the contact surface geometry~\cite{GelSight2009,GelSight2011}. The GelSight sensor consists of three components: (1) soft silicone gel following the shape of the detected object, (2) color LEDs illuminating the deformed membrane and (3) a camera for capturing images. The three-color LEDs illuminate the gel from different angles. Since each surface normal corresponds to a unique color, the color image captured by the camera can be directly used to reconstruct the depth map of the contact surface by looking up a calibrated color-surface normal table. 

The GelSight tactile sensor has already been successfully utilized in various tasks. Li \textit{et al}.~\cite{GelSightTexture} used the GelSight sensor to recognize 40 classes of different surface textures. Jia \textit{et al}.~\cite{GelSightLump} demonstrated that the GelSight sensor outperformed humans in detecting lumps in soft media, indicating the possible application of this sensor on diagnosis of breast cancer. 
To apply the GelSight sensor to robots, Li \textit{et al}.~\cite{GelSightUSB} designed a fingertip GelSight tactile sensor that highly reduced the volume of the GelSight sensor. The sensor was equipped on Baxter robot hand and completed a USB insertion task. Yuan \textit{et al}.~\cite{GelSightShear} further improved the sensor by adding markers on the gel surface. By analyzing the marker motion, the GelSight sensor can be used to sense normal, shear, torsional load on the contact surface and even detect incipient slip. GelSight sensor was also utilized to detect hardness of contact objects based on the analysis of gel deformation and marker displacement in~\cite{GelSightIROS16,GelSightHardness2}.

However, there are two main problems with the GelSight fingertip sensor designed by Li \textit{et al}.~\cite{GelSightUSB} (Li's sensor). First, the sensor can only reconstruct a coarse 3D map of the contact surface. It used acrylic optical guides to illuminate the gel surface, resulting non-uniform illumination on the contact surface. The non-uniform illumination caused errors in the estimated surface normal. In addition, a semi-specular reflective surface was required in Li's sensor to amplify the reflection, but the photometric stereo employed to estimate the 3D height map for the GelSight sensor, does not work well on semi-specular surfaces. Secondly, the fabrication process of this sensor is complicated. Every component of the sensor should be adjusted accurately by hand to ensure that the sensor works properly. The arduous fabrication of the sensor severely restricted its application. 

\subsection{Slip detection}   

Slip detection plays a vital role in performing a successful grasp or manipulation. People have long been trying to develop tactile sensors to sense slip and secure grasp via the detection of physical signals related to slip, like the ratio of shear force to normal force, vibration, acceleration. A recent review of different slip-detection technologies is given in~\cite{slipreview2013}.
Howe and Cutkosky~\cite{howe1989slip} proposed that sensing object acceleration in the hand is the core part of detecting slip or incipient slip, and they made a tactile sensor prototype with an accelerator under the soft rubber surface to detect the slip or incipient slip state of objects.
Holweg~\textit{et al}.~\cite{holweg1996slip} proposed that in the incipient slip stage, there were fluctuations of the contact force between the object and the soft sensor surface. They made a rubber tactile sensor model that predicts the slip by analyzing the frequency of the normal contact force.
Another example is Melchiorri's work~\cite{melchiorri2000slip}, which measured the normal and shear components of the contact force using a force sensor, and compared their ratio to the frictional coefficient of the surface to predict slip. 
Su \textit{et al}.~\cite{biotachslip2015} used a silicone pressure sensor to detect slip by monitoring the sudden change in tangential force and vibration in the normal pressure. 
Ajoudani \textit{et al}.~\cite{IIT2016slipcontrol} built a grasp control system with the slip-detection feature and used the ratio of the shear force to the normal force as an indicator of the likeliness of slip.

Yuan \textit{et al}.~\cite{GelSightShear} proposed that the GelSight sensor could detect slip by analyzing the sensor's contact condition. Their experiments demonstrated that slip started from the peripheral area of the contact surface, where the sensor surface had a relatively smaller displacement. The difference in sensor surface displacement was measured by tracking the motion of the black markers on the sensor. However, they did not perform real-time robot grasping experiments, and only tested objects with flat surface or little surface textures, which guaranteed the contact surface to be large enough to detect the motion of markers.

\section{DESIGN AND FABRICATION OF NEW GELSIGHT SENSOR} 
To improve the accuracy of the 3D height map reconstruction and simplify the fabrication process, here we propose a new design of the GelSight tactile sensor which can use a Lambertian surface instead of a semi-specular surface, optimize the illumination uniformity on the sensing surface, and utilize more standardized and 3D printable framework.

\begin{figure}[t]
    \centering
		\includegraphics[scale=.30]{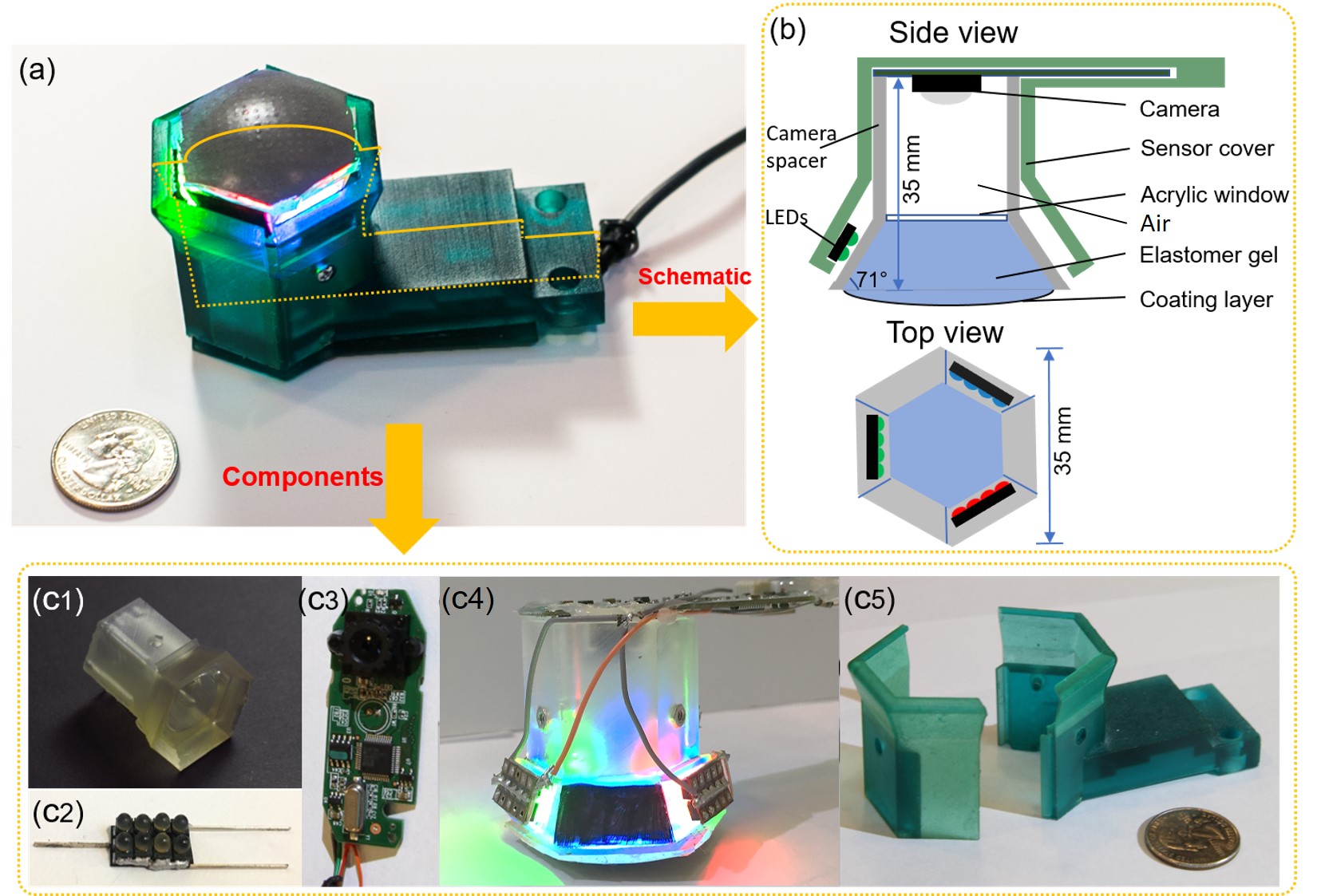}
      \caption{New design of GelSight tactile sensor. (a) The new GelSight sensor. (b) Schematics of the design. (c1)-(c5) The components of the new sensor: 3D printed sensor frame, $2 \times 4$  LED array, Logitech C310 webcam without cover, the prototype of the sensor and  3D printed sensor cover}
      \label{fig:sensor}
\end{figure}

We approach the new design from the following aspects. Firstly, we particularly choose LEDs (Osram Opto Semiconductor Standard LEDs - SMD) with a small collimating lens in front. The lens has a 30$^{\circ}$ viewing angle and efficiently collects and collimates the emitted light from the LED. The LEDs are tightly arranged in a $2 \times 4$ array as shown in Figure~\ref{fig:sensor}(c2). We use three LED arrays with three different colors: red, green, and blue. 

Secondly, we design a hexagonal plastic tray as shown in Figure~\ref{fig:sensor}(c1). This semitransparent tray is produced by a 3D printer (Formlab2) with clear resin. LED arrays are mounted to the every other side of the tray with super glue, as shown in the top view in Figure~\ref{fig:sensor}(b). The mounts of the LED arrays, as noted in Figure~\ref{fig:sensor}(b) is tilted to 71$^{\circ}$ to the sensor surface for illuminating the whole sensing surface. Because of the rotational symmetry, the illuminations of R, G, B near the center of the sensing surface are of equal intensity. The large tilt angles of the illumination also generate a large variance in the reflection of the sensor surface regarding to different surface normals, favoring a more discriminatable surface normal measurement. The semitransparent tray also homogenize the light of the LED array while allows a high transmission.

Owing to the bright and uniform illumination achieved in the new design, we can replace the semi-specular surface with a Lambertian surface. The silicone gel labeled with markers is coated with a thin membrane of aluminum powder mixed in silicone base, which ensures Lambertian reflection. Compared to gel with semi-specular coating, the silicon gel used here is much eaisr to make and the coating material is non-toxic. We design the reflection surface to be dome-shaped, which outperforms the flat reflection surface in terms of the reflection uniformity. The dome-shaped gel surface also makes robotic applications easier. 

For the gel support, a transparent hexagonal-shape acrylic window is laser cut and inserted as a supporting plate. We fill the hollow tray with same silicone materiel to match the refractive index of the silicone gel used for sensing. It eliminates the reflection from the silicone gel, which may otherwise attenuate light and introduce artifacts for imaging. 

For the imaging part, a camera (Logitech C310 in Figure~\ref{fig:sensor}(c3)) is mounted on the top of the hexagonal tray. The camera is 35 mm away from the sensor, allowing a large field of view. Figure~\ref{fig:sensor}(c4) shows the assembled prototype of the sensor according to the design described above. 

In order to increase the durability of the sensor, we further design a plastic protection cover (Figure~\ref{fig:sensor}(c5)) for the prototype of the sensor. The cover is produced by the same 3D printer with a tough material to ensure its robustness. In addition, the handle of the sensor frame is designed to fit the WSG parallel gripper, which makes the sensor easier to be easy to use for many robotic tasks. 

The new design, as shown in Figure~\ref{fig:sensor}(a), has a compact structure. The frame of the sensor is 3D printable, and the illumination parts can be easily glued to the mounts on the surface of the frame. The procedure is standardized and the sensor can be manufactured without any specific skills. The Solidworks file for the sensor frame and cover can be downloaded from \url{http://people.csail.mit.edu/yuan_wz/GelSightData/GelSight_part_172.zip}.

\begin{figure*}[t]
	\centering
 	   \includegraphics[width=\textwidth,height=4.5cm]{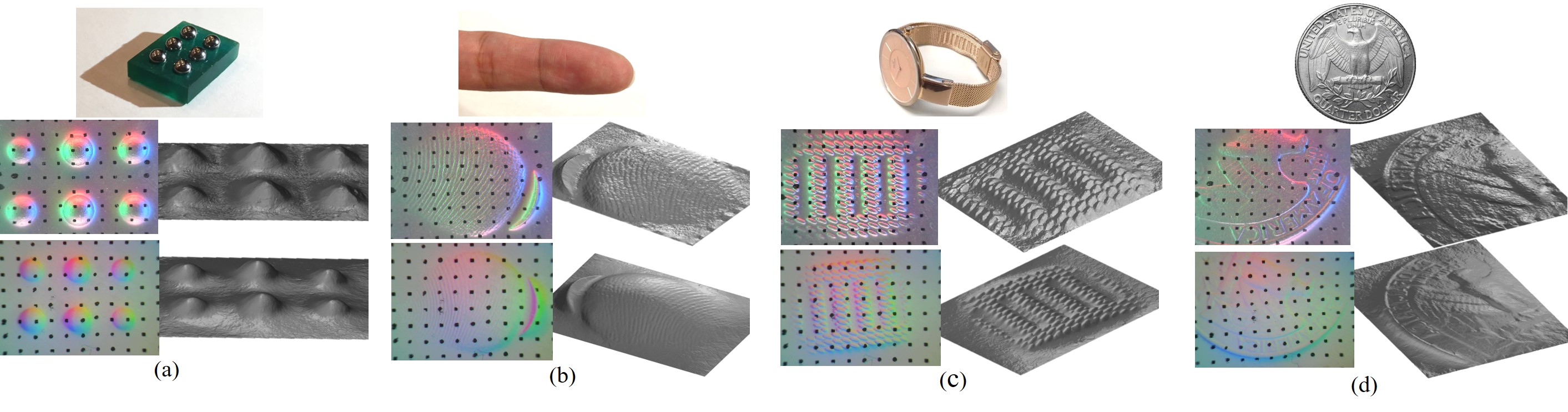}
    \caption{GelSight images of ball array(a), human finger(b), watch chain(c), Quarter(d) and the corresponding reconstructed depth image from new sensor(lower row) and Li's sensor(upper row)}
    \label{fig:3d recovery}
\end{figure*}
Before using the sensor for reconstructing the 3D height map of the contact surface, we calibrate a lookup table that maps R, G, B values to surface normals. The calibration process is performed by pressing a ball with the diameter of 3.96 mm against the surface of the elastomer gel in an arbitrary position. The color value change induced by the distorted gel surfaces is recorded by the camera. The surface normal of each pixel in the image can be calculated according to the diameter of the ball. Afterwards, a lookup table mapping R,G,B values to surface normals for the specific area is automatic generated by the program. To eliminate the noise from the spatial variance of the illumination, this process is repeated at different positions on the surface. The final lookup table is averaged over all tables.

\section{EVALUATION OF GEOMETRY MEASUREMENT WITH GELSIGHT}
To evaluate the performance of the new sensor, we compare Li's sensor and the new sensor  from four aspects: gradient versus color change, mapping accuracy of the lookup table, spatial illumination variance, and quality of 3D shape reconstruction. 

\subsection{Gradient vs. Color Change}
\label{chpt:Color}

\begin{figure}
\centering{
	\includegraphics[scale=0.4]{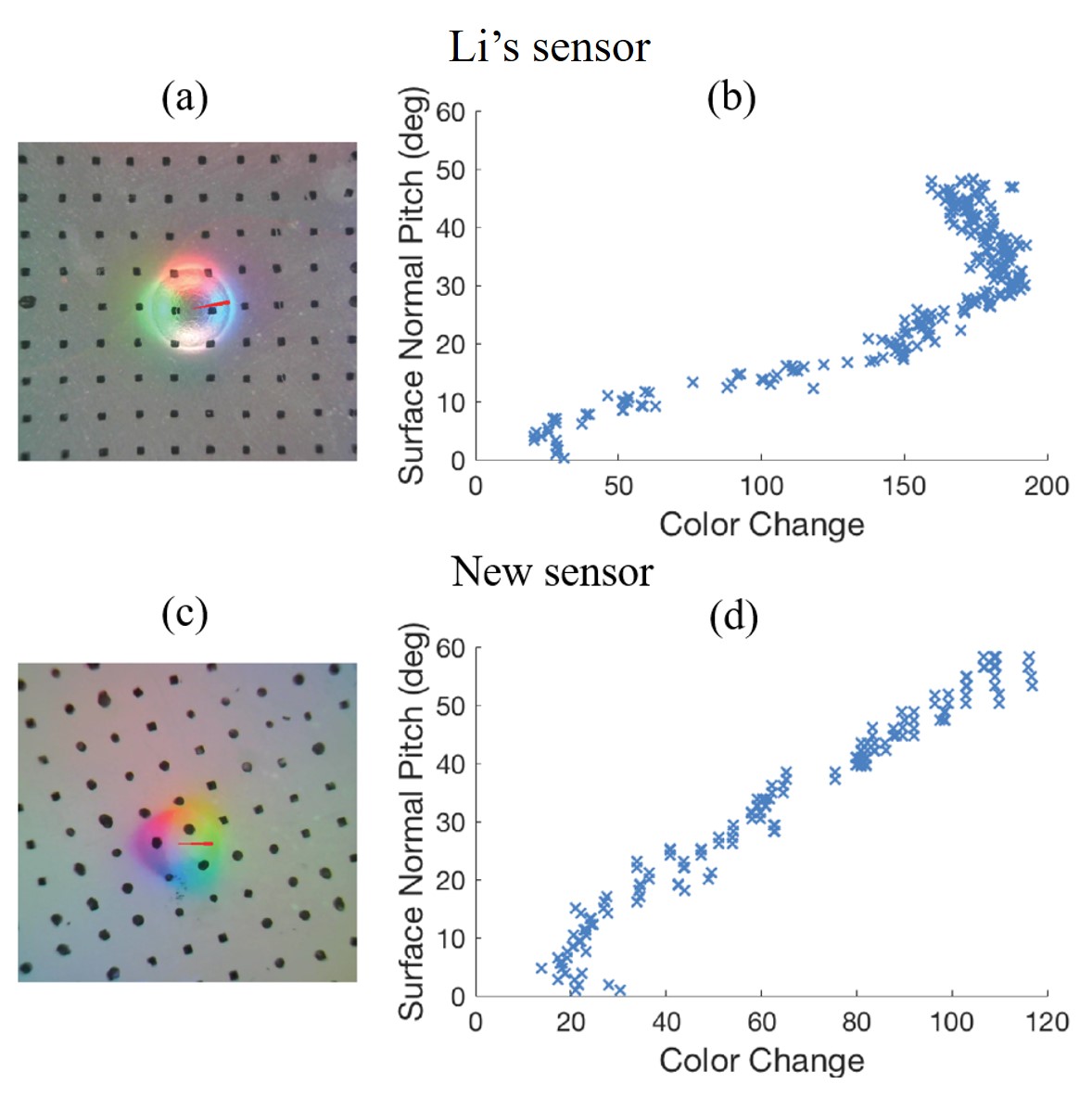}
}
\caption{The color change comparison over the surface normal pitch angle. We take the example of a calibration sphere being pressed on the sensors' surfaces, and compare the change of the GelSight image color over the areas of different surface normals, as shown in the red areas in the left figures. The plots shows, for Li's sensor, the color change is larger upon contact area, but when the surface normal pitch reaches around 30 degree, the color can not well represent the slope change; while for the new sensor, the color change is less obvious, but the linearity is better, within a larger range of slope angles.}
\label{fig:Eval_ColorContrast}
\end{figure}

Since we use color values to infer surface normals, a preferable design should perform a one-to-one mapping between surface normals and color values. We calibrated Li's sensor and the new sensor and choose a pair of calibration images for quantitative analysis. The image of Li's sensor, as shown in Figure~\ref{fig:Eval_ColorContrast}(a), features a rapid color change from the edge of the ball to the center. We inspect the color change with surface normals along the radial direction as denoted by the red arrow in Figure~\ref{fig:Eval_ColorContrast}(a). The color change as a function of the surface normal, as shown in Figure~\ref{fig:Eval_ColorContrast}(b), is approximately linear when the surface normal pitch angle varies from 5 to 25 degrees. Above 30 degrees, the corresponding color change decreases with a higher surface normal pitch angle, indicating that a single color change value is mapped to two or more very different surface normals. The ambiguity prohibits an effective inference from the color values to the surface normals. We implement the same measurement for the new sensor. As shown in Figure~\ref{fig:Eval_ColorContrast}(c), the color varies smoothly with surface normals. The function of the color change and the surface normal is linear from 5 to 60 degrees. We did not measure surface normals larger than 60 degrees because the stiffness of the gel prevented a perfect contact. From the above study, the new sensor presents an improved linear mapping between surface normals and the color change. 

\subsection{Mapping accuracy of the lookup table} 
\begin{figure}
\centering{
\includegraphics[scale=0.44]{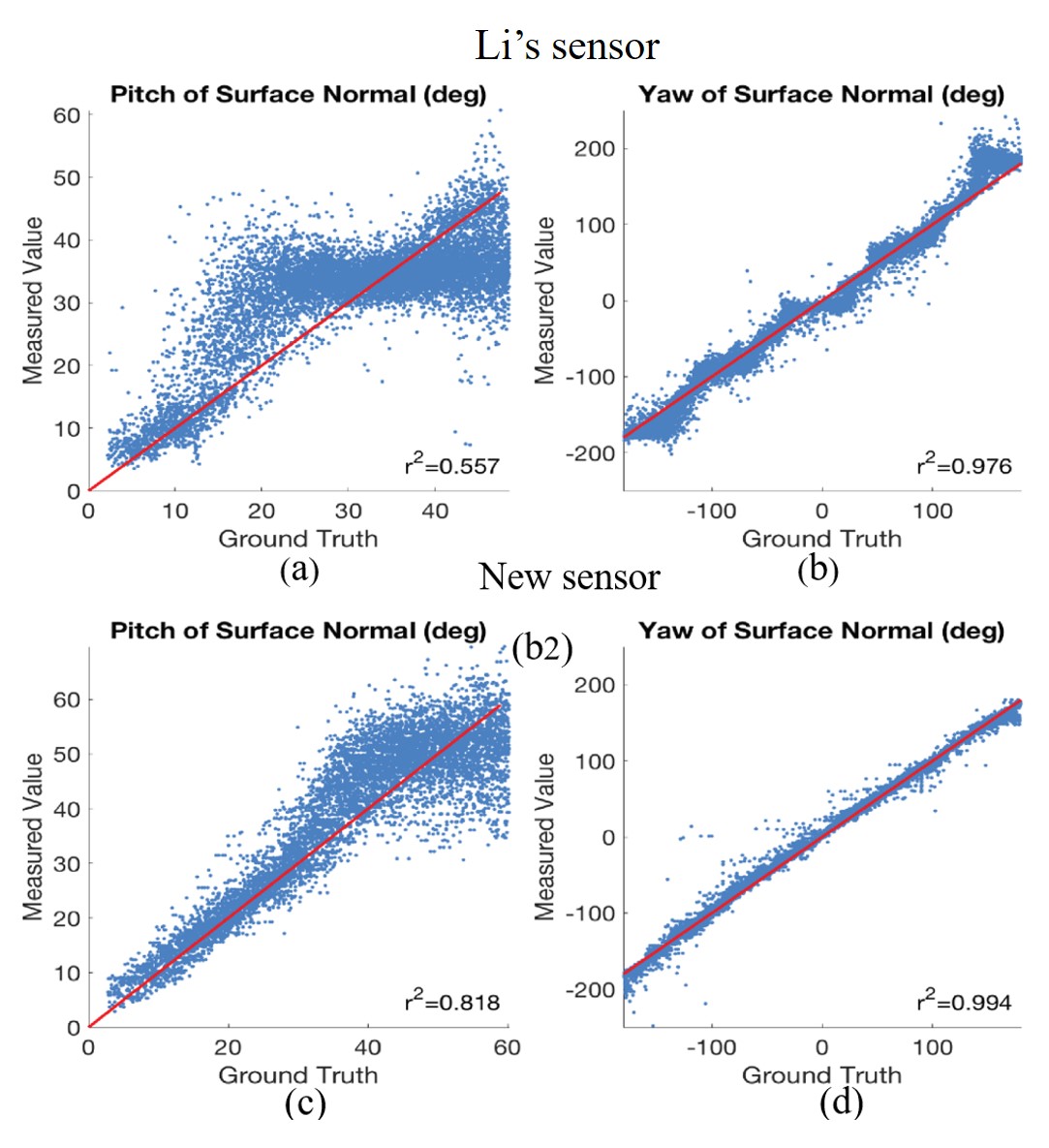}
}
\caption{Comparison of the measured surface normal angles and the ground truth. The plots are based on the images of Li's sensor and new sensor shown in Figure~\ref{fig:Eval_ColorContrast}, when a small sphere is pressed against the sensor surface. We compare the pitch angle and the yaw angles of the pixels within the contact area, and find that the new sensor measures the surface normal more precisely.}
\label{fig:Eval_Lookuptable}
\end{figure}

To evaluate the accuracy of the surface normal measurement, we compared the estimated value with its ground truth of a standard ball. Since both the pitch angle and the yaw angle are indispensable to identify a surface normal, we quantify the two parameters for the Li' sensor and new sensor. Figure~\ref{fig:Eval_Lookuptable} shows the measured surface normal pitch angles vs. the ground truth and measured yaw angles vs. ground truth  for Li's sensor [(a), (b)] and new sensor [(c), (d)]. The dense blue dots represent the data in the lookup table while the red line shows the case of 100\% accuracy. In Figure~\ref{fig:Eval_Lookuptable}(a), the blue dots lay on the red line when the pitch angle varies 5 to 20 degrees. They blow up or drop below the red line as the pitch angle becomes larger than 20 degrees. The $R^2$ is estimated at 0.557, which implies that Li's sensor is not able to retrieve accurate surface normal pitch angles in that range. In contrast, as shown in Figure~\ref{fig:Eval_Lookuptable}(c) all the data here is distributed around the red line and the $R^2$ is improved to 0.818. The new sensor outperforms Li's sensor in terms of retrieving pitch angles. For the surface normal yaw angle, both Li's sensor and new sensor achieve very high $R^2$ values. The new sensor still exceeds, revealing a stronger capability of reconstructing proper yaw angles.

\subsection{Spatial difference of the illumination}  
\begin{figure}
\centering{
\includegraphics[scale=0.3]{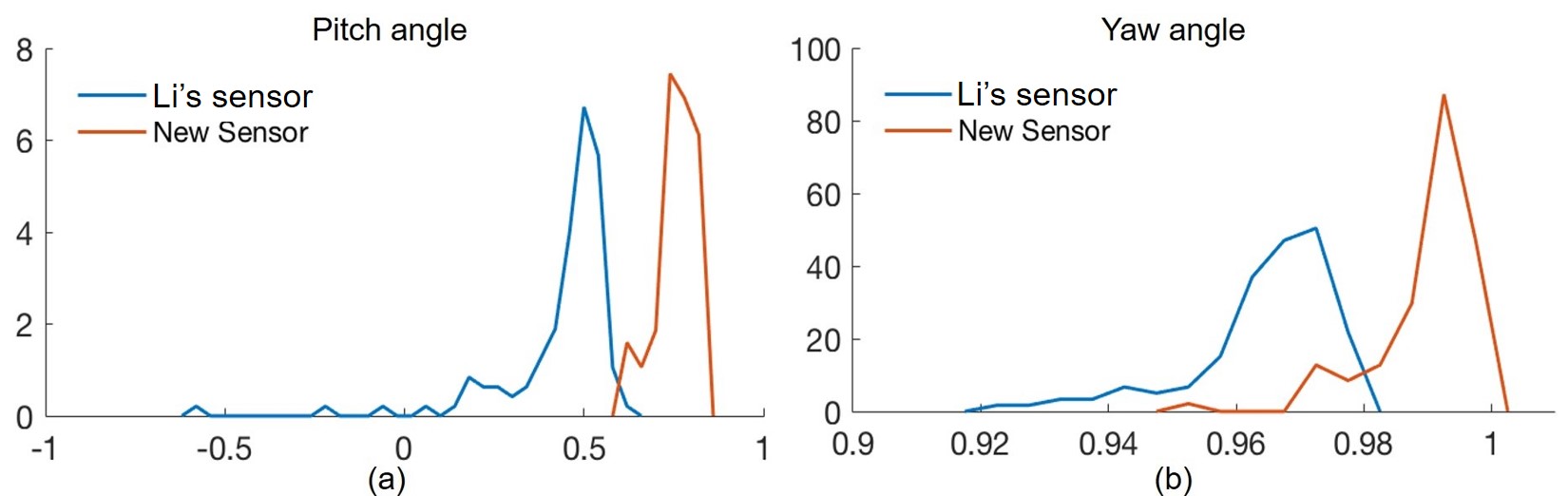}
}
\caption{Probability distribution of $R^2$ of measured surface normals' (a) pitch angle and (b) yaw angle over all the calibration images, on both Li's sensor and new sensor. There are 119 images for Li's sensor and 94 images for the new sensor, where the calibration sphere is pressed at arbitrary locations on the sensors.}
\label{fig:eval_r2Prob}
\end{figure}

Non-uniform illumination results in the variance of the surface normal measurements. 
To quantify the measurement error from this factor, we calculated the $R^2$ of the surface normal pitch angle and the yaw angle at different positions on the sensor surface. The probability distribution of $R^2$ for the pitch angle is plotted in Fig.~\ref{fig:eval_r2Prob}(a). For the new sensor, the peak of the distribution, corresponding to the majority of the $R^2$ values, is larger than the best $R^2$ value obtained of Li's sensor. Moreover, the lowest $R^2$ value of the new sensor is almost equal to the highest $R^2$ of Li's sensor. It is also clear that Li's sensor shows a long tail in the distribution, implying inaccurate predictions of surface normals at certain positions. This tail near low $R^2$ values is, however, absent in the new sensor. The high average $R^2$ value of the new sensor enables a precise measurement of surface normals. The sharp cut-off of the $R^2$ distribution of the new sensor confirms that there is no blind spot on the sensing surface. We can draw a similar conclusion from the $R^2$ value distribution of the yaw angles in Fig.~\ref{fig:eval_r2Prob}(b). The dedicated design of the illumination in the new sensor empowers a more precise measurement of surface normals. 

\subsection{Quality of 3D shape reconstruction}
After the quantitative analysis discussed above, we directly compare the reconstructed 3D images of miscellaneous objects in Figure~\ref{fig:3d recovery}. We tested ball arrays (a), a human finger (b), a watch chain (c), and a quarter coin (d). The depth map recovered by the new sensor (lower panel) is smooth without abrupt changes of the surface normal. The surface recovered by Li's sensor (upper panel) is grainy with a low signal to noise ratio. For example, the depth map of the ball array recovered by the new sensor preserves the shape of a sphere and smooth surface. Contrastingly, the image attained with Li's sensor is deformed as cones with rough surface. Because of the non-uniform illumination, the depth map of the 4 balls on the left and right side shows degraded quality than that of the ones in center. Objects with fine features reveal a higher resolution of the new sensor. In Figure~\ref{fig:3d recovery}(d), with the new sensor (lower panel), the feather of the eagle on the quarter coin looks clear and separable while in the image acquired by Li's sensor, the feather is buried in the noise. Similar observations can be made from the images of the human finger and the watch chain. The
comparison of the 3D maps confirms the improvement of
the new sensor from an intuitive perspective.

\section{SLIP DETECTION WITH NEW GELSIGHT SENSOR}
We use the new GelSight sensor on a robot gripper for grasp tasks, and try to predict slip or incipient slip during grasping.
The GelSight sensor detects slip from 3 major clues: the relative displacement between the objects and the sensor surface, the shear displacement distribution of the markers on the sensor surface, and the change in the contact area. 
Both translational and rotational slips are considered.

For objects with obvious textures, GelSight can precisely recognize the texture location, and track the object's motion according to the texture motion; the movement of the sensor surface is indicated by the black markers on the surface. A relative movement between the object texture and the markers, either translational or rotational movement, indicates the occurrence of slip. 
For objects with large curvature and smooth surface, like a coke can, GelSight can detect the objects' shape, but not their precise movement, because the geometry of the contact surface changes little during the slip. For these objects, we use the method described in~\cite{GelSightShear}: slip occurs starting from the peripheral contact area, which makes the sensor surface in the peripheral area have less displacement compared to the central area. The difference can be inferred by comparing the motions of different markers.
In any case, if there is a significant decrease in the contact area, we know a severe slip has occurred. To make the slip detection efficient so that it can be applied to real-time robot grasp tasks, we simplified the algorithm according to the experimental data.

\begin{figure}
	\centering{
		\includegraphics[scale=0.3]{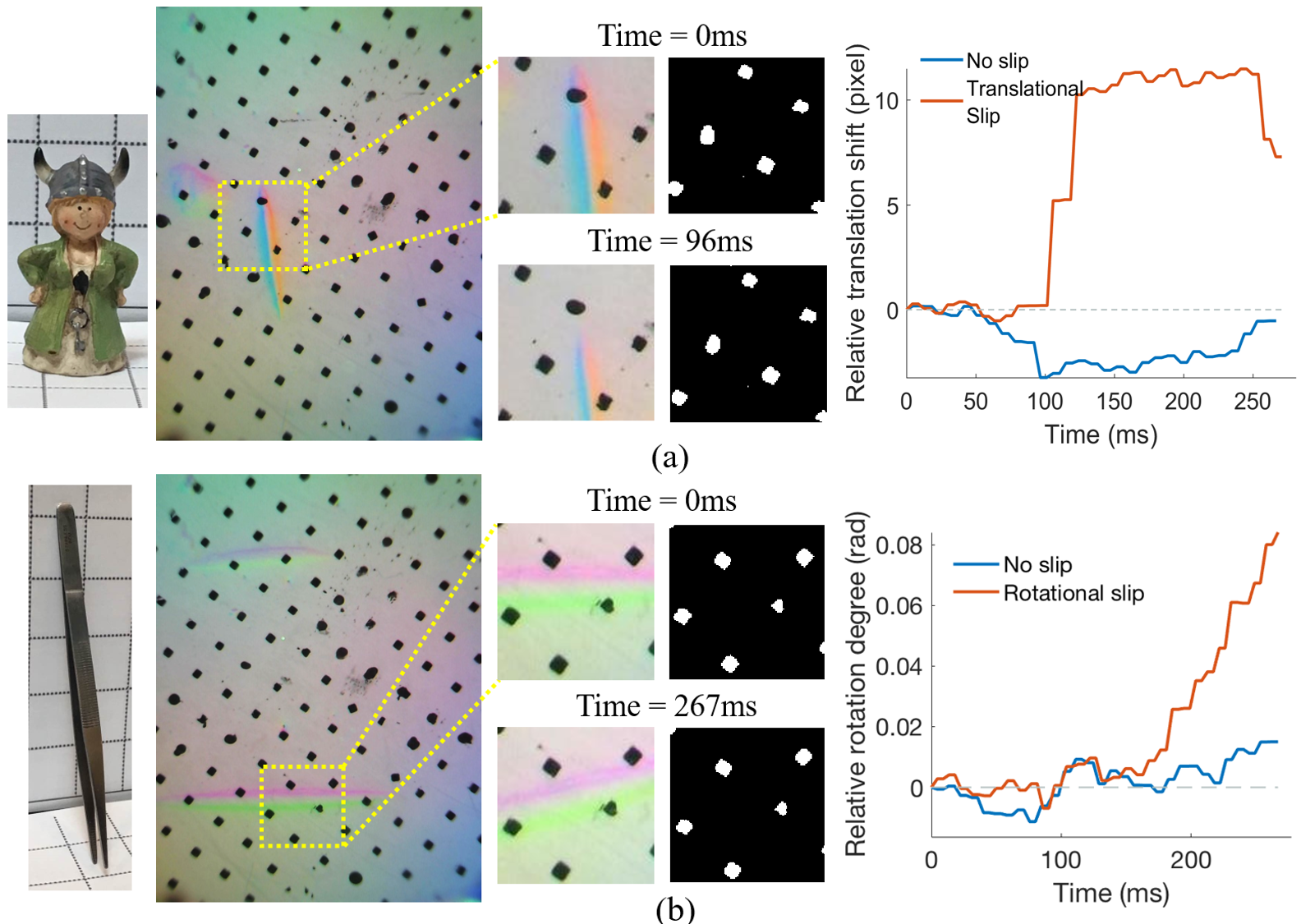}}
	\caption{Translational and rotational slip detection based on geometry and markers, when the object have a obvious texture. We crop a patch in the GelSight image, and compute the translation/rotation of both the color patterns and the markers in the patch, and compare their differences. We consider slip to occur when the differences are large. The plot compared the relative translation/rotation of the color texture and the markers, when slip occurs and not. } 
	\label{fig:TextureDetect}
\end{figure}
\textbf{Slip detection: measuring the relative displacement between object texture and markers} We calculate the translation and rotation of the markers and the object texture between two adjacent frames, and compare the motion differences between the markers and texture. If there is a significant difference that exceeds the threshold, or the accumulated relative translation or rotation is large, we consider slip is happening.

In practice, we select a $120 \times 120$ window on the frame, centered at the pixel with the largest difference in the color intensity, corresponding to a contact area with the largest curvature. This window is usually in the middle of the contact area if the object has a textured surface. For objects with larger surface curvatures, like a pen, this window is typically in the border area. To check the texture motion, we use the gray-scaled image of the window based on the color intensity change. Figure~\ref{fig:TextureDetect} shows an example for the relative translation and rotation between the objects and markers.

\begin{figure}
	\centering{
		\includegraphics[scale=0.22]{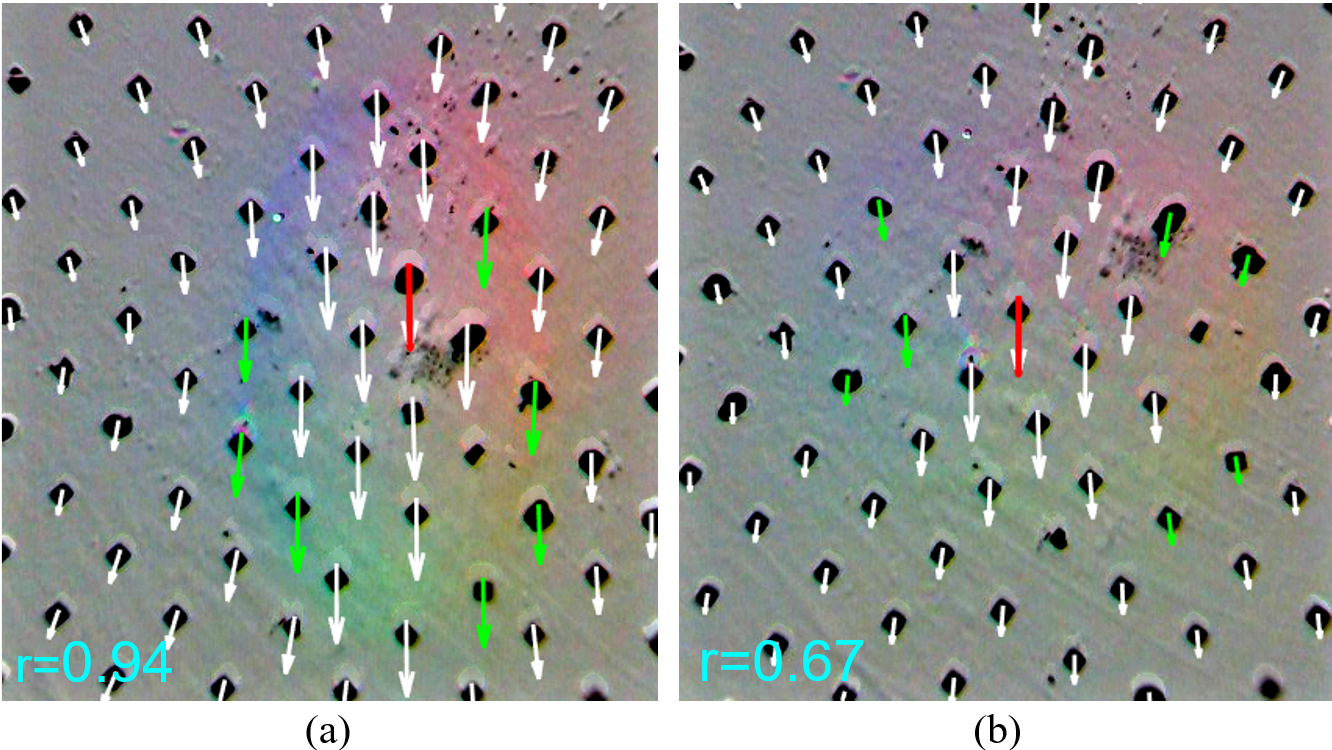}
	}
	\caption{
		The motion of GelSight markers when gripper lifts (a) a plastic cylindrical bottle and (b) a wood cubic block, where slip occurs in (b), and the markers in the contact area have more varied motion. The backgrounds in the pictures are the color intensity change of GelSight images. 
		To detect the slip through marker motion, we compare the moving distance of the 'approximate peripheral area markers $\mathbf{v}_p$' (shown as the green arrows), and the maximum marker motions $\|\mathbf{v}_i\|_{\operatorname{max}}$ (shown as the red arrows). When their ratio $r$ is lower than a threshold, we predict slip occurs.}
	\label{fig:markerSlip}
\end{figure}
\textbf{Slip detection: tracking the marker motion distribution in the contact area} For all the motion vectors $\mathbf{v}_i$ of the markers, we measured the maximum moving distance $ \|\mathbf{v}_i\|_{\operatorname{max}}$, which is usually from some markers in the middle of the contact area. We also measured the moving distances of the markers in the peripheral area $\mathbf{v}_p$, and calculated the following ratio

\begin{equation}
	r=\frac{\|\mathbf{v}_p\|_{\operatorname{max}}}{\|\mathbf{v}_i\|_{\operatorname{max}}}
\end{equation}

If $r$ is larger than a threshold (we selected 0.8 in the experiments), we consider the slip is occurring or going to occur soon, and the robot should execute some protection procedure. 

To identify markers located in the peripheral areas, we compute the change in the average pixel intensity of the nearby area for each marker and chose the ones with the largest intensity changes, which corresponds to the largest gradient of the surface geometry (See Section~\ref{chpt:Color} and Figure~\ref{fig:Eval_ColorContrast}) and matches the border of the contact area. We select 8 markers (might be repetitive), the surrounding areas of which have the largest color change across different color channels, and pick the moving vector with the largest norm ($\|\mathbf{v}_p\|_{\operatorname{max}}$) among these 8 markers to ensure a safe selection. Figure~\ref{fig:markerSlip} shows two examples of how the markers are chosen.

This method also works, although less perfectly, for detecting rotational slip. When there is torque on the contact surface, an appropriate indicator is to compare the rotational angles of all the markers: in the incipient slip or slip stage, the markers in the peripheral area have smaller rotational angles compared to ones in the center. However, the computation of flow center is time-consuming, so we only use the norms of the motion vectors of different markers to detect rotational slip. 



\textbf{Contact detection}
We intend to make the pressing force of the robot gripper small and controllable. 
Additionally, during the lifting process, a decrease in the contact area indicates a severe slip is occurring. We detecte the contact with a simplified method that compares the color intensity of current image with that of the image obtained when nothing contacts the sensor surface. For objects with a smooth and flat surface, we also calculate the overall motion of the markers to track the changes in the normal force~\cite{GelSightShear}.

\section{Experimental result on robotic slip detection}

\subsection{Experimental setup}
\label{chpt:Setup}

We conduct robot grasp experiment with a robot system composed of a UR5 6DOF arm and a WSG 50 parallel gripper. The GelSight sensor is mounted on the gripper as one of its fingers, as shown in Figure~\ref{fig:sensor in gripper}(a). The reach of the robot arm is 850mm, and the repeatability is $\pm 0.1$mm. The gripper has a maximum opening width of 110mm, which is then reduced to 80mm when a GelSight sensor is mounted. The minimum closing speed of the gripper is 5mm/s. 

\begin{figure}
\centering{
	\includegraphics[width= 3.45in]{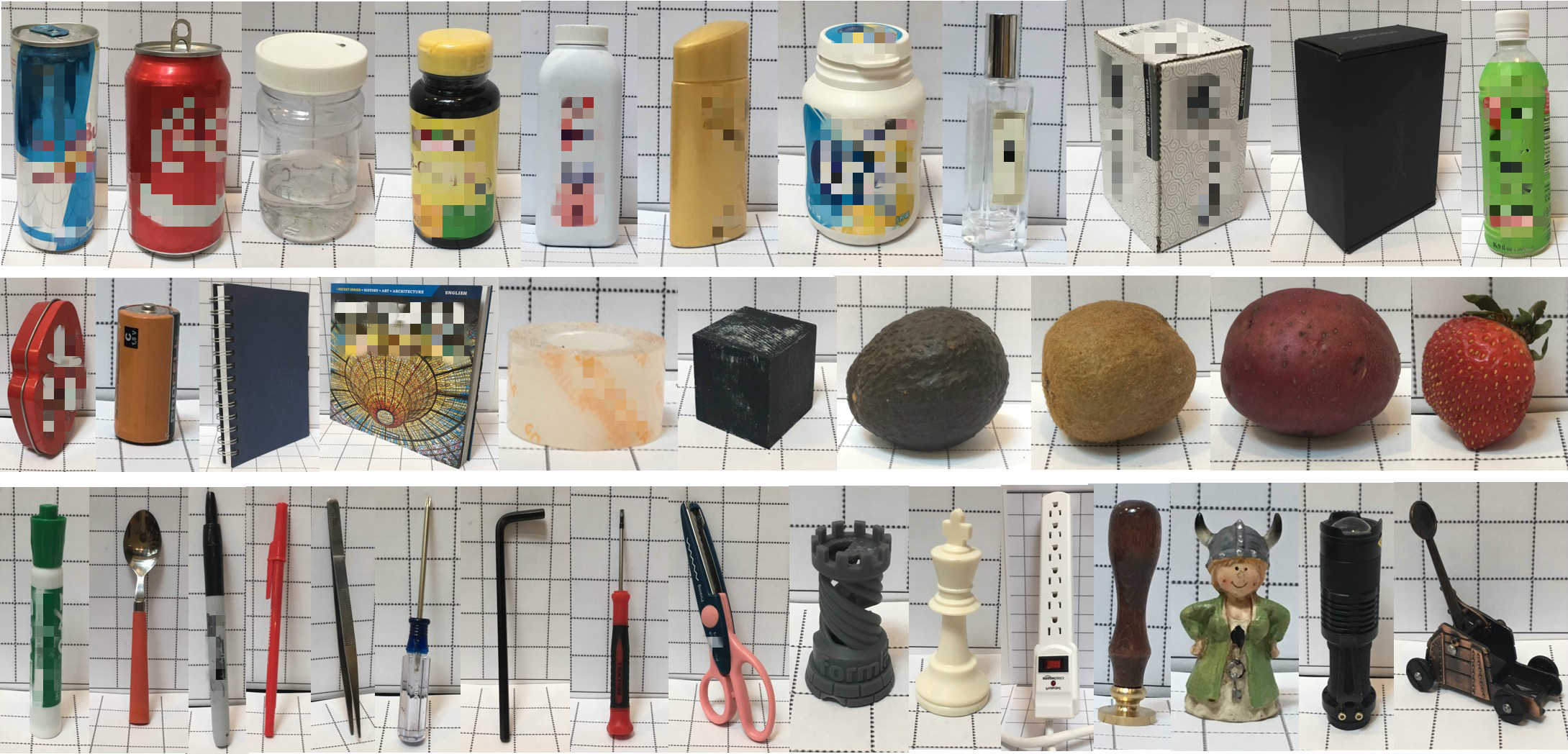}
}
\caption{The 37 objects for the grasp experiment. They are commonly seen objects in everyday life, with different sizes, shapes and materials. }
\label{fig:objects}
\end{figure}

We perform the grasp experiments on 37 natural objects that are commonly seen in everyday life, as shown in Figure~\ref{fig:objects}. The objects are different in size, shape, material and surface texture.
In the experiment, the gripper stopped when it is in the proximity of the target object, and slowly approach the object to eventually grasp. The GelSight signal informs whether the finger has contacted the objects and the force is large enough. After grasping, the robot lifts the object slowly for 3cm, and then stops in the midair. During the lifting, the algorithm detects whether slip is occurring. 

We also use the GelSight feedback on slip detection to achieve a safe grasp. In this experiment, if slip is detected during the lift process, the robot stops and puts the object down, releases it, and then re-grasp it with a larger threshold for contact detection. The robot keeps this loop until the grasp is considered safe.

\subsection{Slip prediction}

In the first experiment, we use the robot to grasp the 37 objects described in Section~\ref{chpt:Setup} and Figure~\ref{fig:objects}, and slowly lift them. Each object is grasped 7 to 10 times, with different contact threshold for grasp, which means the gripper forces are different. For most objects, we try to equal the number of cases when the grasp is successful or slip occurs; for a small portion objects, because they are very light or the surface is too smooth, there are only the cases for successful grasp or slip.

We record the GelSight video during the grasp and lift process, and measure ``whether slip occurred'' by human observation. We put the tests in three groups according to the results: successful cases, when there is little relative motion between the objects and the gripper, and the gripper grasped the objects firmly; failure cases, when either significant translational or rotational slip occurred and the grasp failed; border cases, when the gripper barely lifted the objects, but not tightly so that the object easily dropped under small external interference, or even slip has occurred before equilibrium. For the border case, we consider it fine to measure either `slip' or `no slip', but we record the rate when GelSight measures `no slip'.
The results for the three groups are shown in Table~\ref{Table:Res}. 

\begin{table}
	\caption{Experimental Result on Slip Prediction}
	\label{Table:Res}
	\begin{center}
		\begin{tabular}{c|c|c|c}
			\hline
			& Successful cases & Slip cases & Border cases \\
			\hline
			\hline
			Sample Number &147 & 116 & 52 \\
			\hline
			Correct Measure &  79\% & 84\%  & 60\%\\
			\hline
		\end{tabular}
	\end{center}
\end{table}

A typical cause for the prediction failure occurs when the object has a flat and smooth surface, and the gripping force is very small. In this case the gripper simply slide along the gripper surface. As the shear force is so small, the noise prevents the system to detect the slip through marker motion analysis, while there is not enough textures to detect the slip. This situation causes 28\% of the failure in the slip cases. Another typical case is that the marker measurement method fails when there was a significant rotational slip when grasping a flat object. This situation causes 28\% of the failure in the slip cases, and it can be prevented if a more thorough measurement on the marker rotational movement is conducted. 
\subsection{Grasp control with slip detection}

Our second experiment is to grasp objects in a close loop with the feedback from the GelSight slip detection. Similar to the first experiment, the robot gripper grasps and lifts the objects, but if slip is detected by GelSight, it suspends and releases the object, and then re-grasps it on the same position using a higher contact threshold, i.e., a larger gripping force.
We test on 33 objects from Figure~\ref{fig:objects}, while ignore the other 4 that the robot could not lift. Each object is grasped for 3 times and we collect 99 grasp cases. 
We count the number of cases that the gripper finally grasped the objects stably. 

The gripping force is controlled by the GelSight signal, based on the surface geometry change or the marker displacement. For each objects, a different threshold is required for a stable grasp, but we set the initial threshold to the same value -- a very small one indicating a bare contact with the object. We increase the threshold by 1.2 times for each trial. 

Out of the 99 grasp experiments, the robot successfully grasped the objects for 88 times, i. e., a success rate of 89\%. In each grasp experiment, the target object is grasped for 1 to 6 times, and the average is 2.3 times. When GelSight detected slips, mostly human can clearly see the slip occuring. 
In the failure cases, the gripping forces are so small that there are not enough information from the contact area for the sensor to measure slip effectively.

\section{CONCLUSIONS}

In this work, we designed a new Fingertip GelSight tactile sensor. We demonstrated that a precise 3D shape of the contact surface can be reconstructed with this new sensor. Compared to Li's sensor~\cite{GelSightUSB}, the new sensor features lower reconstruction errors and smaller spatial variances. The fabrication process is standardized and easy to implement. We performed slip detection with the sensor in a grasping experiment. For  37 objects tested  in this work, the new sensor can detect both translational and rotational slip during grasping without any prior knowledge of the objects.  The implementation of the new sensor assists safe manipulation tasks. The new Fingertip GelSight sensor can find various applications such as safe grasping, object recognition, and hardness estimation. 

\addtolength{\textheight}{-1cm}   





\section*{ACKNOWLEDGMENT}

The work is supported by Toyota Research Institute. We thank Wen Xiong, Dongying Shen, Changchen Chen, Abhijit Biswas for revising the manuscript. We thank Shaoxiong Wang for helping set up robot arm.


\bibliographystyle{IEEEtran}
\bibliography{Ref}

\end{document}